\documentclass[10pt,twocolumn,letterpaper]{article}
\usepackage[margin=0.75in,columnsep=0.25in]{geometry}
\usepackage{newtxtext}
\usepackage{graphicx}
\usepackage{booktabs}
\usepackage{amsmath,amssymb,amsthm}
\usepackage{newtxmath}
\usepackage[numbers,sort&compress]{natbib}
\usepackage{xcolor}
\usepackage{caption}
\usepackage[colorlinks=true,linkcolor=blue,citecolor=blue,urlcolor=blue]{hyperref}
\usepackage{microtype}
\usepackage{algorithm}
\usepackage{algpseudocode}
\newcommand{\eps}{\varepsilon}
\renewcommand{\phi}{\varphi}
\newcommand{\best}[1]{\textbf{#1}}
\newcommand{\wraw}{w}
\newcommand{\wema}{\bar w}

\definecolor{propbg}{gray}{0.92}
\newcounter{prop}
\newsavebox{\propbin}
\newenvironment{propx}[1]{%
  \par\medskip\noindent\refstepcounter{prop}%
  \begin{lrbox}{\propbin}\begin{minipage}{\dimexpr\linewidth-2\fboxsep\relax}%
  \vspace{5pt}%
  \textbf{\itshape Proposition \theprop\ (}#1\textbf{\itshape ).}\ \itshape\ignorespaces
}{%
  \vspace{5pt}\end{minipage}\end{lrbox}%
  \noindent\colorbox{propbg}{\usebox{\propbin}}\par\medskip
}

\title{Catastrophic Overfitting is Weight Drift\\the Training Loss Cannot See}

\author{Anonymous CVPR submission}
\date{}

\begin{document}
\raggedbottom

\twocolumn[{%
\begin{center}
{\LARGE\bfseries Catastrophic Overfitting is Weight Drift the Training Loss Cannot See\par}
\vspace{8pt}
{\large Chao Pan\par}
\vspace{2pt}
{\small Department of Computer Science and Engineering,\\
Southern University of Science and Technology, Shenzhen 518055, China\par}
{\small \texttt{11930665@mail.sustech.edu.cn}\par}
\end{center}
\vspace{2pt} }]

\noindent{\small\bfseries Abstract.} {\small Fast adversarial training (FAT) makes robust training affordable by replacing the multi-step attack with a single-step one, but it suffers from catastrophic overfitting (CO): partway through training, robustness to multi-step attacks collapses to near zero while the single-step attack used for training keeps being repelled. We show that CO is a quiet escape of the weights. Across the event the network's clean behavior quietly improves, and moving from the weights just before CO to those just after it changes the training objective only marginally while destroying most of the robustness. The training objective is therefore nearly blind to the displacement it causes, so it exerts no force to undo it. The exponential moving average (EMA) of the weights, which FAT protocols already maintain, does not accumulate this drift and stays robust through the entire event. This observation suggests a nearly free fix: when validation robustness drops, reset the weights to their EMA. We build this guard on top of a coupling-based base method designed so that its average stays inside the robust region. The resulting training removes CO in every one of our runs and attains the strongest AutoAttack accuracy among fast methods on all three benchmark datasets, including a clear margin on CIFAR-100, where it also clears the multi-step reference at roughly half its cost. Code is available at \url{https://github.com/fzjcdt/co-weight-drift}.\par}

\vspace{2pt}

\begin{figure}[!t]
\centering \includegraphics[width=\linewidth]{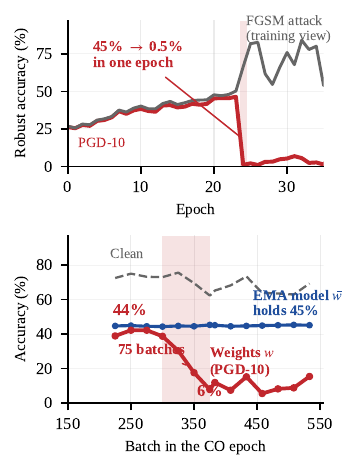}
\caption{\textbf{Catastrophic overfitting (CO) at two timescales.} \emph{Top:} at epoch resolution, CO in vanilla FGSM training is a cliff without warning; validation PGD-10 robustness falls from $45\%$ to $0.5\%$ in a single epoch (shaded), while the FGSM attack used for training keeps being repelled, so nothing inside the training loop looks wrong. \emph{Bottom:} the same method retrained with checkpoints every 25 batches. The cliff is $75$ batches wide, the EMA model $\bar w$ (blue) holds its robustness throughout, and clean accuracy keeps rising. Our guard resets the weights to this EMA when validation robustness dips.}
\label{fig:teaser}
\end{figure}

\section{Introduction} \label{sec:intro}

Adversarial training \cite{madry2018pgd} is the most reliable known defense against adversarial examples, but it repeats a multi-step attack for every batch, an order of magnitude more computation than ordinary training. Fast adversarial training (FAT) \cite{goodfellow2014explaining,shafahi2019adversarial,wong2020fast} removes this bottleneck by training against a single-step attack. It works well, with one strange exception.

The exception is \emph{catastrophic overfitting} (CO) \cite{wong2020fast,andriushchenko2020understanding,kim2021understanding}. Partway through an otherwise ordinary run, robustness to multi-step attacks collapses to near zero within a single epoch. The single-step FGSM attack used for training continues to be repelled, so from the inside of the training loop nothing appears to be wrong, and the run finishes with a model that is robust against nothing except the attack it trained on. Because CO is silent and cannot be predicted in advance, the strongest published fast methods are chosen as much for their luck as for their design.

Many fixes have been proposed: randomized initialization \cite{wong2020fast} and gradient-alignment regularizers \cite{andriushchenko2020understanding}, modified attacks and initializations \cite{kim2021understanding,nfgsm,pan2024fgsmuap}, bi-level formulations \cite{wang2024pco}, and others \cite{he2023selffitting,zhao2025atss}. They help empirically, but they all act on the attack or on the loss, and none of them answers the mechanistic question: \textbf{where in weight space has the CO model gone, and why is the optimization that was robust an epoch earlier unable to find its way back?}

Part of the reason nobody has answered it is that ordinary training logs cannot see the event: at epoch resolution CO is a cliff with no consistent precursor. So we rebuilt the instrument. We retrained the classic victim, vanilla FGSM adversarial training, and every 100 batches saved \emph{two} checkpoints: the training weights and their exponential moving average (EMA), an object the protocol already maintains. This turns a one-epoch mystery into a short film.

Three findings came out of watching that film (Fig.~\ref{fig:teaser}). First, \emph{CO barely touches the clean function}: through the event, clean accuracy rises while PGD-10 robustness falls from $44\%$ to $11\%$. Robustness leaves before accuracy notices. Second, \emph{the direction of travel is invisible to the training objective}: interpolating from the pre-CO EMA to the CO weights changes the training loss by a few percent while destroying more than two thirds of the robustness, so there is no restoring force along the direction that matters. Third, \emph{the weight average resists the drift}: the EMA keeps its robustness through the entire event, and in the benchmark's released runs its own CO, when it happens at all, arrives many epochs later or never.

These three facts point to a fix that is not another regularizer. If training cannot correct the drift, but the trajectory's own average does not accumulate it, the natural move is to pull the weights back toward their EMA when the drift shows up. We build this idea in its simplest form: a \emph{guard} that resets the weights to their EMA when validation robustness drops by more than a threshold, sitting on top of a strong coupling-based base method designed so that its EMA holds the robust region. On CIFAR-10 the unguarded base collapses in half of its runs; the guarded version removes the coin flip entirely and reaches the best fast-method AutoAttack accuracy we are aware of on all three benchmark datasets, including a clear gain on Tiny-ImageNet.

\noindent\textbf{Contributions.} (1)~A measured mechanism: CO is a drift of the weights along directions the training objective can barely see, established by paired raw/EMA tracking at 100-batch resolution and by line searches along the drift direction (Figs.~\ref{fig:pair},~\ref{fig:linesearch}). (2)~Its dynamics: a short warning window that no epoch-level observable can see, ending in gradient masking (Fig.~\ref{fig:micro}), and a staged, margin-ordered sweep through the evaluation set. (3)~A fix derived from the mechanism: anchoring the weights to their own EMA, with a threshold chosen inside a measured gap and a detector-free control showing that the rescue comes from the anchor itself. (4)~Results: the strongest fast-method AutoAttack accuracy on CIFAR-10 ($50.45{\pm}0.20$, 4 seeds, no CO), CIFAR-100 ($27.15{\pm}0.04$, $+0.51$ over the strongest fast method and $+0.25$ over the multi-step reference), and Tiny-ImageNet ($+0.59$); the benchmark's second strong-attack suite (CR) confirms the ordering on all three.

\section{Related Work} \label{sec:related}

\paragraph{Catastrophic overfitting.} CO was identified with single-step adversarial training in \cite{wong2020fast}. Later work attributed it to the degradation of the single-step attack and proposed randomized initialization \cite{wong2020fast} and gradient-alignment regularizers \cite{andriushchenko2020understanding}, attack and initialization modifications \cite{kim2021understanding,nfgsm,pan2024fgsmuap}, bi-level formulations \cite{wang2024pco}, and self-fitting analyses \cite{he2023selffitting}. All of these act on the attack or on the loss, and each new fix has been followed by the failure reappearing in a new form. \textbf{Our analysis acts on the weights instead: it locates the CO solution relative to the robust one and shows that the direction between them is nearly flat for the training objective}, which explains both why attack-side interventions help without explaining and why the failure keeps returning.

\paragraph{Fast adversarial training.} Single-step methods with consistency or spectral couplings define the current FAT frontier: FGSM-PGI \cite{fgsmpgi}, N-FGSM \cite{nfgsm}, LIET \cite{pan2025liet}, NuAT \cite{sriramanan2024nuat}, GAT \cite{gat}, SSAT \cite{kim2021understanding}, and the 21-method benchmark \cite{pan2026benchmark} whose protocol and released results we reuse throughout. \textbf{We differ in what the couplings are for: previous methods use them to prevent CO, while we use them to keep the weight average inside the robust region so that a reset has somewhere to return to}, and we separate that role from the guard that performs the reset. All baseline numbers in this paper are the benchmark's released results; nothing is retrained for comparison.

\paragraph{Weight averaging.} EMA and SWA \cite{polyak1992acceleration,izmailov2018averaging} average weights after or alongside training; Lookahead \cite{zhang2019lookahead} unconditionally pulls fast weights toward slow weights; EWC \cite{kirkpatrick2017overcoming} anchors to old \emph{tasks} by Fisher importance. \textbf{We use the average as a live control target, gated by a robustness measurement rather than applied unconditionally}, and the difference matters: an always-on anchor at moderate pull strength destroys training, because it also pulls back the component the optimizer is legitimately optimizing (Appendix~\ref{sec:dose}).

\section{Anatomy of Catastrophic Overfitting} \label{sec:anatomy}

\subsection{What we trained and what we saved} \label{sec:instrument} We retrain the classic victim, vanilla FGSM adversarial training (FGSM-AT in the benchmark of \cite{pan2026benchmark}), under the benchmark's exact protocol (ResNet18, CIFAR-10, $\eps{=}8/255$; Appendix~\ref{app:proto}). Every 100 batches we save two checkpoints together: the raw training weights $w$ and their EMA $\wema$ (decay $0.9995$, the value the benchmark protocol itself uses). All evaluations in this section recalibrate BatchNorm statistics first; without this step the entire effect disappears (Appendix~\ref{app:proto}). On this setup CO lands at epochs 19--21 across three seeds, and we track both checkpoints through the window.

\begin{figure}[t]
\centering \includegraphics[width=\linewidth]{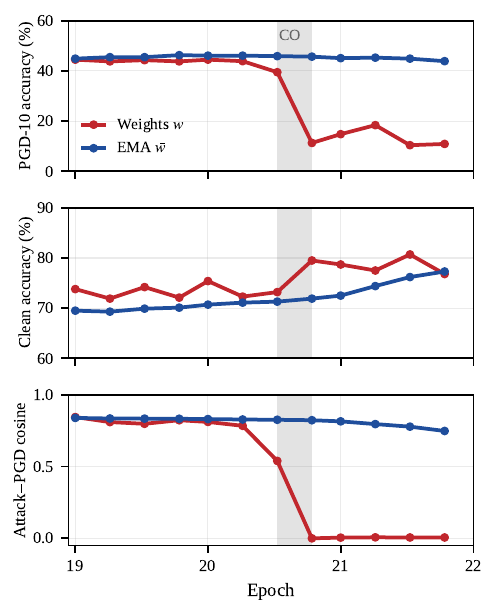}
\caption{What collapses and what does not. Every 100 batches through the CO window we evaluate the raw weights $w$ (red) and their EMA $\bar w$ (blue), BatchNorm recalibrated; the gray band marks the 100 batches in which CO happens. Top: the weights' PGD-10 accuracy falls from $44\%$ to $11\%$ while the EMA holds $44$--$46\%$. Middle: over the same interval the clean accuracy of the weights \emph{rises} from $73\%$ to $80\%$. Bottom: the cosine between the training-attack direction and a PGD direction falls from $0.82$ to $0.00$ in the weights and stays high in the EMA.}
\label{fig:pair}
\end{figure}

\subsection{Robustness leaves before accuracy notices} \label{sec:def} Fig.~\ref{fig:pair} shows the paired timeline. Between the last healthy checkpoint and the first CO one, 100 batches apart, the raw weights change dramatically for robustness and hardly at all for everything the training loop can observe. Clean accuracy moves from $73\%$ to $80\%$. The cosine between the training-attack direction and a PGD-10 direction falls from $0.82$ to $0.00$. Over the same interval, PGD-10 robustness falls from $44\%$ to $11\%$. \textbf{Robustness leaves before accuracy notices, and nothing the optimizer monitors moves in a way that would call the drift back.}

The EMA tells the other half of the story. Evaluated on the same grid, it holds PGD-10 robustness at $44$--$46\%$ and its gradient cosine at $0.75$--$0.83$ for the entire event. This is a general property, not an artifact of one run: across the benchmark's released per-epoch curves, the EMA's own CO arrives 6--24 epochs after the raw CO in 4 of the 11 CIFAR-10 CO runs and never within training in the other 7 (Fig.~\ref{fig:lag} in the appendix). The drift reaches the average only slowly.

\subsection{A flat path for training, a cliff for robustness} \label{sec:linesearch} To make ``invisible'' precise, let $\mathcal L$ be the training objective (the empirical risk under the training attack) and $R$ the PGD-10 robustness. We call a direction $v$ \emph{blind} if moving along it leaves $\mathcal L$ nearly unchanged while destroying $R$:
\begin{equation}
|\mathcal L(w{+}\alpha v)-\mathcal L(w)|\le \eps\alpha,\quad
|R(w{+}\alpha v)-R(w)|\ge \delta\alpha,
\label{eq:def}
\end{equation}
with $\delta/\eps\gg 1$. The term is operational: it means flatness by measurement, normalized relative to each function's own level, and makes no claim about null spaces of Hessians or NTK kernels.

\begin{figure*}[t]
\centering \includegraphics[width=0.72\textwidth]{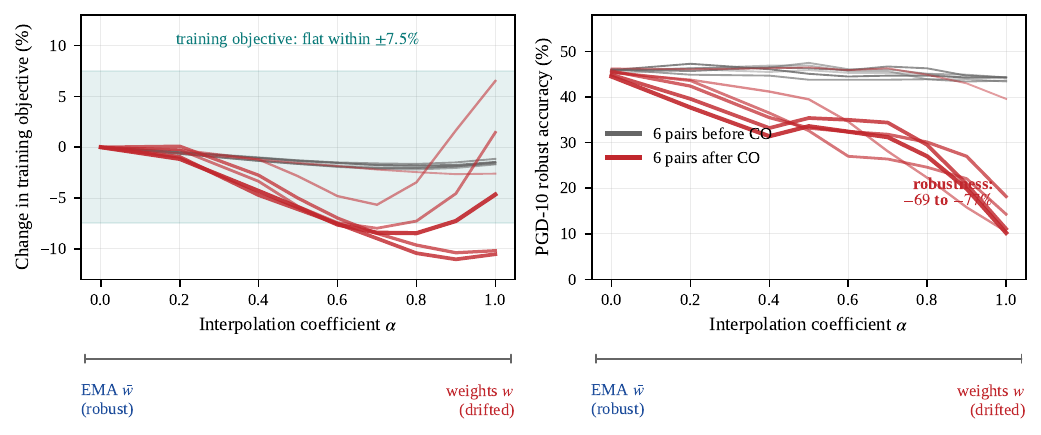}
\caption{Moving along the drift direction, $w(\alpha)=\bar w+\alpha\,(w-\bar w)$, for 12 (EMA, weights) checkpoint pairs sampled every 100 batches through the CO window, with the training attack recomputed at each $\alpha$; each curve runs from the EMA at $\alpha{=}0$ (left) to the weights at $\alpha{=}1$ (right). Gray: the 6 pairs from before CO. Red, darkening with time: the 6 pairs from after CO. Left: the training objective varies within $\pm 7.5\%$ of its level along every path, and the CO endpoint is not a better training loss. Right: along the paths after CO, PGD-10 loses $69$--$77\%$.}
\label{fig:linesearch}
\end{figure*}

We then measure how blind the actual drift is (Fig.~\ref{fig:linesearch}). We take 12 (EMA, raw) checkpoint pairs through the CO window, one every 100 batches, and for each pair interpolate $w(\alpha)=\wema+\alpha(\wraw-\wema)$ from the EMA ($\alpha{=}0$) to the weights ($\alpha{=}1$), recomputing the attack at every $\alpha$. The training objective varies within $\pm 7.5\%$ of its level along the entire path, and the CO endpoint is not even an improvement. Over the same path, robustness loses $69$--$77\%$, with a knee at $\alpha\approx0.6$--$0.8$: the sensitivity gap along this single direction is roughly $10\times$. \textbf{There is no barrier between the robust and the CO solutions; they are connected by a path that is flat for training and a cliff for robustness.} The model did not find a better minimum; it wandered sideways off a plateau.

\subsection{Why training cannot undo it} \label{sec:props} The measurements above compress into three short statements about the dynamics. We state them informally, with assumptions inline; each is backed by a measurement in the preceding sections.

\begin{propx}{no restoring force along blind directions}
\label{prop:force} If $v$ satisfies Eq.~\eqref{eq:def} at $w$, the expected SGD update along $v$ is $\mathbb E\langle\Delta w,v\rangle=-\eta\langle\nabla\mathcal L,v\rangle=O(\eta\eps)$: the dynamics exert essentially no force on the blind component, which instead drifts under gradient noise.
\end{propx}

\begin{propx}{the time-average compresses the drift}
\label{prop:avg} If the blind displacement is stationary with no persistent bias, its running average satisfies $\mathbb E\|\bar v_T\|^2=O(\sigma^2/T)$ while the raw displacement grows as $\mathbb E\|v_T\|^2=O(\sigma^2 T)$.
\end{propx}

\begin{propx}{resetting to the average is safe}
\label{prop:reset} Replacing $w$ by $\wema$ keeps the visible component of the trajectory and discards the accumulated blind displacement; by Eq.~\eqref{eq:def} the replacement perturbs $\mathcal L$ by at most $O(\eps)$ (measured: $+6$ points of clean accuracy) while restoring $R$ toward its pre-CO level.
\end{propx}

\noindent Together they say: training cannot correct the drift, the average does not accumulate it, and resetting to the average is cheap for the training objective. That is exactly the fix we build in Sec.~\ref{sec:method}.

Blind directions are not exotic. A perturbation $v$ of the last layer that increases the logit margin $z_y-\max_{k\ne y}z_k$ without changing any argmax preserves clean predictions exactly; the softmax gradient of the cross-entropy decays exponentially in the margin, so $\langle\nabla\mathcal L,v\rangle\to0$, and the saturated logit surface creates a flat region in which few-step attacks stall while longer attacks still find the true boundary. This is the gradient masking we measure next. The assumption of no persistent bias (Proposition~\ref{prop:avg}) draws the method's applicability boundary: it holds for coupled FAT methods, whose EMA holds the robust region through the event, and fails for vanilla FGSM training, whose biased drift drags the average along; anchoring indeed fails there (Appendix~\ref{sec:scope}).

\subsection{The last warning window} \label{sec:micro} 
\begin{figure}[t]
\centering \includegraphics[width=\linewidth]{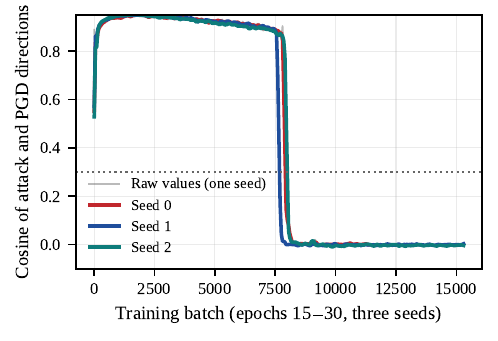}
\caption{CO at batch resolution. The cosine between the training-attack direction and a PGD direction (three seeds, smoothed; faint trace: raw values of one seed) decays over 200--300 batches before the robustness cliff. At epoch resolution the same event is a featureless cliff.}
\label{fig:micro}
\end{figure}

Why has this geometry gone unnoticed? Because at epoch resolution the event broadcasts nothing reliable: none of the epoch-level statistics we tried fires consistently before the cliff. At 50-batch resolution the picture changes (Fig.~\ref{fig:micro}). The cosine between the training-attack direction and a PGD direction decays from $0.79$ to $0.32$ over 200--300 batches, a soft ramp rather than a cliff, and only then does PGD-10 robustness fall. This ramp is the only warning window that exists, and every per-epoch observable misses it.

After the ramp, the model fakes robustness to weak attacks. PGD-3 (three steps of $2/255$) reports robust accuracy \emph{rising} to $86\%$ while PGD-10 destroys the model, and the FGSM probe accuracy rises as well. This is textbook gradient masking \cite{athalye2018obfuscated}, now measured batch-by-batch as the \emph{end state} of CO: the training attack has become a no-op, so the method trains on effectively clean data while believing it trains adversarially. Two practical consequences follow: weak iterative attacks are unreliable probes of CO, and evaluation protocols that mix attack budgets can mis-rank models across the transition (precursor statistics in Appendix~\ref{app:analyses}).

\subsection{The cliff, sample by sample} \label{sec:death} The aggregate cliff is 75 batches wide, which invites the picture of a simultaneous global event. The samples tell a different story. Re-evaluating the checkpoints of the window on the 1{,}000 validation images, with the attack seed fixed so that every checkpoint sees the identical PGD-10 initialization, we call a sample \emph{dead} at the first checkpoint where it stays fragile for three consecutive checkpoints. Deaths are staged: a cohort of $16\%$ of the initially robust samples dies $100$ or more batches before the median, so the aggregate cliff is the visible end of a slower sweep (Fig.~\ref{fig:death} in the appendix). The death order is written in the healthy model: early deaths have a smaller clean logit margin at the window's first checkpoint, and the order is a property of the samples rather than of one training path, since an independently trained replica reproduces it (cross-replica Spearman $\rho{=}0.48$; full statistics in Appendix~\ref{app:analyses}). CO sweeps the evaluation set in a largely preserved order, starting from the samples whose correct-class margin was smallest all along. This structure does not yield an earlier trigger; its value is to say what the drift destroys first, and to ground the cliff shape in a sweep rather than a simultaneous global event.

\section{Pulling the Weights Back to Their Average} \label{sec:method}

The three propositions point to a simple rule rather than a new regularizer. The drift cannot be corrected by training. The average does not accumulate it. Resetting is nearly free for the training objective. So: watch for the drift, and when it appears, reset the weights to their own EMA (Eq.~\ref{eq:reset}). Before the control itself, we specify the training it sits on top of, because the base method's properties are what make anchoring possible at all.

\subsection{A base method built for anchoring} \label{sec:base}

The guard needs something to guard: a training configuration whose EMA actually holds a robust region. The configuration is dictated by what the coupling terms must measure. Write $x_a=x+\delta_a$ for the adversarial example of a batch. The perturbation $\delta_a$ is what the training attack itself produces at that step: it starts from a fresh random initialization $\delta_0$ (random half-sign at $0.5\eps$, following NuAT \cite{sriramanan2024nuat}), takes one sign step of the cross-entropy loss evaluated at $x{+}\delta_0$, and is clipped to the threat model, $\|\delta_a\|_\infty\le\eps$ (Algorithm~\ref{alg:guard}, lines~3--4). It is an input-space pattern, recomputed for every batch, not a weight-space direction. Writing the clean and adversarial logits as $z_c=f(x)$ and $z_a=f(x{+}\delta_a)$, the coupling $\|z_c-z_a\|_2^2$ is a finite-difference estimate of the directional derivative of $f$ along the adversarial displacement, anchored at the clean input $x$. The anchor matters: $x$ lies on the data manifold and does not move with the weights, so the estimate is made where the labels live and is statistically independent of the current parameters. Anchoring elsewhere, for example at the attack's initialization point $x{+}\delta_0$, makes the penalty self-referential: it rewards flattening exactly the gradient direction the attack uses, a masking feedback loop that we measured and that destroys the robust region (Appendix~\ref{app:seeds}). The clean branch is not regularization; it is information.

The full objective couples the two branches at two scales:
\begin{equation}
\mathcal{L} \;=\; \mathrm{CE}_{\mathrm{ls}}(z_c, y)
\;+\; \frac{\gamma(t)}{B}\,\big\| Z_c - Z_a \big\|_*
\;+\; m(t)\,\big\| z_c - z_a \big\|_2^2 ,
\label{eq:loss}
\end{equation}
where $Z_c, Z_a$ stack the batch's clean and adversarial logits, $\|\cdot\|_*$ is the nuclear norm (sum of singular values), $B$ is the batch size, $\mathrm{CE}_{\mathrm{ls}}$ is label-smoothed cross-entropy, and $\gamma(t), m(t)$ ramp linearly over training (values per dataset in Sec.~\ref{sec:exp}). The two coupling terms act at different scales and cover each other's failure modes. \emph{The MSE term} follows the per-example consistency idea of FGSM-PGI \cite{fgsmpgi}, which penalizes the distance between the current adversarial prediction and a prior-guided initialization, but anchors the penalty at the clean input: it aligns each image's clean and adversarial logits, and a setup with only this term is stable but plateaus below the frontier. \emph{The nuclear term}, the idea behind NuAT \cite{sriramanan2024nuat}, acts on the batch as a matrix: it penalizes concentration of the clean-minus-adversarial logit difference into a few directions, and alone it is toxic at the same strength on the harder datasets, exactly where the published nuclear-norm method fails (Appendix~\ref{app:seeds}). \emph{Together} the setup is the strongest fast configuration we measured when it survives. Both terms need the clean logits in the computation graph, so each step carries three forward passes; we also tried to recover the clean branch at two-pass cost and failed in every arrangement we could think of, which we read as the price of true-label supervision on the data manifold (Appendix~\ref{app:seeds}).

Two properties of this setup matter for the rest of the paper. First, \textbf{it is strong but drift-prone}: on CIFAR-10 half of its runs suffer CO, which makes it the honest testbed for the anchor, since a method that never collapsed would need no guard. Second, its drift carries no persistent bias, the assumption of Proposition~\ref{prop:avg}: the coupling terms suppress the biased drift of vanilla FGSM training, which is why its EMA holds the robust region through a CO event. The base and the anchor are designed together; neither works as well alone.

\subsection{The guard: a discrete reset} \label{sec:guard} Every method in the benchmark already evaluates PGD-10 on a held-out validation split every epoch, for checkpoint selection. The guard reads exactly this number for the raw weights. Writing $p_t$ for the epoch-$t$ validation PGD-10 of $w_t$, it maintains the running best and the deficit
\begin{equation}
b^{\star}_t=\max_{s\le t} p_s, \qquad
g_t \;=\; b^{\star}_t - p_t ,
\label{eq:guard}
\end{equation}
and fires the reset
\begin{equation}
w_{t+1} \;=\;
\begin{cases}
\wema_t, & g_t>\theta \ \wedge\ t-t_{\mathrm{fire}}\ge 3
\ \wedge\ F_t<5,\\[2pt]
w_t, & \text{otherwise},
\end{cases}
\label{eq:reset}
\end{equation}
where $t_{\mathrm{fire}}$ is the last reset epoch and $F_t$ the number of resets so far (Algorithm~\ref{alg:guard}). The two bookkeeping constants exist for failure modes, not for normal operation: the cooldown stops the guard from firing every epoch right after a reset (the anchor lands below the pre-CO peak, so the deficit stays positive by construction), and the fire budget is a stop-loss against a sinking anchor (on vanilla FGSM training an uncapped guard fires 68 times in 100 epochs). In our formal runs the bookkeeping does its work quietly: three of the four CIFAR-10 seeds never fire at all, and the fourth fires three times inside its CO window, each dip spaced exactly at the cooldown, recovering to the same level as the others ($50.61$). The cost is zero additional passes: the deployed method is a static network trained and evaluated under the unchanged protocol.

\begin{algorithm}[t]
\caption{Guarded coupled FAT (one training run). Lines~3--4 define
$\delta_a$: a fresh half-sign initialization, one sign step, clipped to
the threat model. The guard reads only the validation PGD-10 the
protocol already computes.}
\label{alg:guard}
\begin{algorithmic}[1]
\State $\bar w \leftarrow w$;\quad $b^{\star} \leftarrow 0$;\quad
$fires \leftarrow 0$ \Comment{$\bar w$: EMA, $\tau{=}0.9995$}
\For{epoch $t = 0, \dots, T{-}1$}
  \For{each minibatch $(x,y)$ of size $B$}
    \State $\delta_0 \sim$ fresh half-sign init at $0.5\eps$;\quad
    $z_0 = f(x{+}\delta_0;\, w)$, \; $z_c = f(x;\, w)$
    \State $\delta_a = \mathrm{clip}\big(\delta_0 + \eps\,
      \mathrm{sign}\,\nabla_{\delta_0}\,\mathrm{CE}(z_0, y)\big)$;\quad
    $z_a = f(x{+}\delta_a;\, w)$
    \State $\mathcal{L} \leftarrow$ Eq.~\ref{eq:loss} with
      $\gamma(t), m(t)$ \Comment{the coupled step}
    \State $w \leftarrow w - \eta\nabla_w \mathcal{L}$;\quad
      $\bar w \leftarrow \tau\,\bar w + (1{-}\tau)\, w$
  \EndFor
  \State $p_w \leftarrow$ validation PGD-10 of $w$
    \Comment{protocol already computes it for selection}
  \State $b^{\star} \leftarrow \max(b^{\star},\, p_w)$
  \If{$b^{\star} - p_w > \theta$ \textbf{ and } $fires < 5$ \textbf{ and } cooldown $\ge 3$ epochs}
    \State $w \leftarrow \bar w$;\quad $fires \leftarrow fires + 1$
    \Comment{reset to the anchor; zero cost}
  \EndIf
\EndFor
\State \textbf{return} weights selected by best validation PGD-10
  (standard protocol)
\end{algorithmic}
\end{algorithm}

\begin{figure*}[t]
\centering \includegraphics[width=0.86\textwidth]{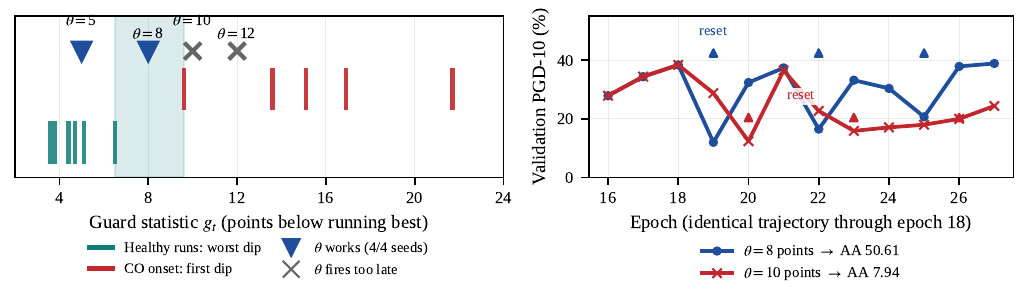}
\caption{Left: choosing the threshold. Over 600 epochs of healthy runs the guard statistic never dips more than $6.5$ points below its best; the first abnormal epoch of every CO dips at least $9.6$. We place $\theta{=}8$ points in the gap. Right: the same seed under two thresholds (logged validation curves; arrows mark guard resets). The trajectories are bit-identical through epoch 18; $\theta{=}8$ catches the first abnormal epoch and recovers (AA $50.61$); $\theta{=}10$ misses it, fires one epoch later on a contaminated anchor, and dies ($7.94$).}
\label{fig:threshold}
\end{figure*}

\subsection{Choosing the threshold} \label{sec:window} The guard's one parameter has a natural value, because healthy and drifting trajectories are well separated (Fig.~\ref{fig:threshold}). Across 600 epochs of healthy guarded runs the deficit $g_t$ never exceeds $6.5$ points, while at the onset of every CO we have observed the first abnormal epoch dips at least $9.6$ points. Any threshold in $(6.5,9.6)$ therefore never fires on healthy trajectories and always fires on the first abnormal epoch; we choose the center, $\theta{=}8$ points. The choice is also forgiving in one direction and sharp in the other, and one controlled experiment shows how sharp: on a seed whose trajectory is identical through epoch 18, $\theta{=}8$ catches the first abnormal epoch and the run finishes at $50.61$ AutoAttack, while $\theta{=}10$ fires one epoch later on an already contaminated anchor and the run dies at $7.94$ (Fig.~\ref{fig:threshold}, right).

Two families of controls pin down that the rescue comes from the anchor itself rather than from the detection, and where anchoring stops helping; both are relegated to the appendix. A detector-free control that resets at a fixed epoch rescues the same seed essentially as well (Sec.~\ref{sec:dose}), and an always-on continuous pull of the weights toward the EMA either sleeps through slow drift or, at higher strength, destroys training (Sec.~\ref{sec:trigger}). Anchoring requires a robust region to anchor to: it cannot save vanilla FGSM training, whose average drifts along (Sec.~\ref{sec:scope}).

\section{Experiments} \label{sec:exp}

\paragraph{Protocol.} We follow the benchmark protocol \cite{pan2026benchmark} exactly and reproduce no baselines: ResNet18 \cite{he2016resnet} on CIFAR-10/100 and PreActResNet18 on Tiny-ImageNet, $\eps{=}8/255$, 100 epochs, per-epoch validation PGD-10 for checkpoint selection, AutoAttack \cite{croce2020autoattack} and the benchmark's constraint-relaxation (CR) attack \cite{pan2025crattack} on the full test set; full details in Appendix~\ref{app:proto}. Our method is the base setup of Eq.~\eqref{eq:loss} plus the guard, with $\gamma_{\max}{=}4.0$, $m_{\max}{=}2.5$ on CIFAR-10, and $\gamma_{\max}{=}1.25$, $m_{\max}{=}4.5$ on CIFAR-100 (the coupling strength must be re-balanced on the harder dataset; Sec.~\ref{sec:nuc}). All $\pm$ values are sample standard deviations over the listed seeds.

\begin{table*}[t]
\centering
\caption{Robust accuracy (\%) under AutoAttack (AA) and the benchmark's constraint-relaxation attack (CR) on all three benchmark datasets, $\eps{=}8/255$, mean$\pm$sample std over the released seeds. Baselines are the benchmark's released results \cite{pan2026benchmark}; nothing is retrained for comparison. Ours: base method plus guard, 4 seeds on CIFAR-10/CIFAR-100 and 3 on Tiny-ImageNet. Bold: best fast method per column. The multi-step rows (PGD-AT, PGD-AT-WA) are shown for reference at several times the fast-method cost.}
\label{tab:main} \small \setlength{\tabcolsep}{4.5pt}
\begin{tabular}{lcccccc}
\toprule
& \multicolumn{2}{c}{CIFAR-10} & \multicolumn{2}{c}{CIFAR-100} & \multicolumn{2}{c}{Tiny-ImageNet} \\
\cmidrule(lr){2-3}\cmidrule(lr){4-5}\cmidrule(lr){6-7}
Method & AA & CR & AA & CR & AA & CR \\
\midrule
\multicolumn{7}{l}{\emph{multi-step (reference)}}\\
PGD-AT \cite{madry2018pgd} & 44.33$\pm$0.08 & 44.50$\pm$0.13 & 21.44$\pm$0.23 & 20.81$\pm$0.30 & 17.04$\pm$0.08 & 16.84$\pm$0.12 \\
PGD-AT-WA \cite{izmailov2018averaging} & 50.49$\pm$0.15 & 50.51$\pm$0.16 & 26.90$\pm$0.17 & 26.87$\pm$0.17 & 19.77$\pm$0.16 & 19.75$\pm$0.13 \\
\midrule
\multicolumn{7}{l}{\emph{fast}}\\
NuAT \cite{sriramanan2024nuat} & 50.10$\pm$0.18 & 50.11$\pm$0.19 & 14.41$\pm$3.29 & 17.66$\pm$0.42 & 10.20$\pm$0.44 & 10.19$\pm$0.44 \\
LIET \cite{pan2025liet} & 49.99$\pm$0.15 & 49.93$\pm$0.15 & 26.64$\pm$0.13 & 26.63$\pm$0.14 & 19.88$\pm$0.09 & 19.86$\pm$0.07 \\
FGSM-PGI \cite{fgsmpgi} & 49.89$\pm$0.08 & 49.89$\pm$0.12 & 26.29$\pm$0.16 & 26.31$\pm$0.12 & 19.55$\pm$0.11 & 19.52$\pm$0.12 \\
FGSM-FUAP \cite{pan2024fgsmuap} & 49.73$\pm$0.12 & 49.71$\pm$0.12 & 25.95$\pm$0.08 & 25.94$\pm$0.05 & 19.10$\pm$0.18 & 19.09$\pm$0.19 \\
FGSM-CUAP \cite{pan2024fgsmuap} & 49.55$\pm$0.22 & 49.54$\pm$0.22 & 25.85$\pm$0.50 & 25.83$\pm$0.49 & 19.19$\pm$0.09 & 19.20$\pm$0.11 \\
FGSM-UAP \cite{pan2024fgsmuap} & 49.44$\pm$0.14 & 49.37$\pm$0.15 & 25.84$\pm$0.24 & 25.82$\pm$0.23 & 18.90$\pm$0.28 & 18.96$\pm$0.21 \\
GAT \cite{gat} & 49.15$\pm$0.15 & 49.15$\pm$0.17 & 22.06$\pm$0.08 & 21.97$\pm$0.07 & 11.33$\pm$0.06 & 11.27$\pm$0.06 \\
FGSM-RS-CS \cite{zhao2023convergesmooth} & 48.64$\pm$0.38 & 48.70$\pm$0.34 & 26.20$\pm$0.26 & 26.20$\pm$0.26 & 18.74$\pm$0.11 & 18.69$\pm$0.11 \\
FGSM-PCO \cite{wang2024pco} & 48.52$\pm$0.32 & 48.48$\pm$0.34 & 19.97$\pm$0.43 & 19.91$\pm$0.39 & 11.24$\pm$0.21 & 11.20$\pm$0.23 \\
GradAlign \cite{andriushchenko2020understanding} & 47.99$\pm$0.62 & 48.02$\pm$0.61 & 25.37$\pm$0.14 & 25.39$\pm$0.15 & 17.87$\pm$0.19 & 17.81$\pm$0.22 \\
N-FGSM \cite{nfgsm} & 47.77$\pm$0.41 & 47.84$\pm$0.41 & 24.86$\pm$0.46 & 24.89$\pm$0.49 & 18.79$\pm$0.08 & 18.76$\pm$0.06 \\
N-AAER \cite{lin2023aaer} & 47.38$\pm$0.18 & 47.48$\pm$0.20 & 22.74$\pm$1.17 & 22.81$\pm$1.18 & 8.74$\pm$11.65 & 8.74$\pm$11.65 \\
ZeroGrad \cite{golgooni2023zerograd} & 45.94$\pm$2.19 & 46.04$\pm$2.20 & 25.00$\pm$0.14 & 24.95$\pm$0.14 & 18.15$\pm$1.11 & 18.10$\pm$1.14 \\
FGSM-MEP-CS \cite{zhao2023convergesmooth} & 45.35$\pm$0.81 & 45.37$\pm$0.78 & 24.86$\pm$0.28 & 24.80$\pm$0.26 & 18.23$\pm$0.19 & 18.13$\pm$0.16 \\
Free-AT \cite{shafahi2019adversarial} & 44.34$\pm$0.17 & 44.32$\pm$0.24 & 21.76$\pm$0.25 & 21.74$\pm$0.27 & 13.51$\pm$0.12 & 13.51$\pm$0.15 \\
FGSM-RS \cite{wong2020fast} & 43.09$\pm$5.33 & 43.14$\pm$5.29 & 21.42$\pm$2.24 & 21.46$\pm$2.24 & 16.33$\pm$0.67 & 16.26$\pm$0.71 \\
SSAT \cite{kim2021understanding} & 38.33$\pm$1.55 & 38.32$\pm$1.53 & 18.83$\pm$0.73 & 18.79$\pm$0.79 & 11.94$\pm$0.12 & 11.92$\pm$0.14 \\
FGSM-AT \cite{goodfellow2014explaining} & 37.28$\pm$0.30 & 37.27$\pm$0.27 & 16.62$\pm$0.54 & 16.58$\pm$0.50 & 14.87$\pm$3.77 & 14.84$\pm$3.75 \\
ELLE \cite{rocamora2024elle} & 32.46$\pm$25.27 & 39.77$\pm$12.77 & 21.21$\pm$2.37 & 21.23$\pm$2.37 & 18.81$\pm$0.21 & 18.80$\pm$0.21 \\
\midrule
\textbf{Base + guard (ours)} & \best{50.45$\pm$0.20} & \best{50.44$\pm$0.16} & \best{27.15$\pm$0.04} & \best{27.20$\pm$0.07} & \best{20.47$\pm$0.11} & \best{20.54$\pm$0.15} \\
\bottomrule
\end{tabular}

\vspace{2pt} \raggedright\scriptsize The unguarded base is not deployable: on CIFAR-10 it matches the guarded row on the two seeds that survive (AutoAttack $50.67{\pm}0.08$) but returns gradient-masked artifacts on the two that collapse, for a 4-seed mean of $29.1$; the guard's effect is exactly the removal of this failure mode. On CIFAR-100 and Tiny-ImageNet the guard never fires; those columns are the re-balanced base method's own performance with the guard carried as zero-cost insurance. Under the iterative attacks of the released suite the ordering is similar (PGD-50: LIET $55.97$, NuAT $54.63$, ours $54.63$); full columns in the supplementary material.
\end{table*}

\paragraph{CIFAR-10.} The base method is the strongest fast configuration we know when it survives, and it survives half the time. \textbf{The guard removes the coin flip: four runs, zero CO, $50.45{\pm}0.20$ AutoAttack}, $+0.46$ over LIET, the strongest published fast method, and $+0.35$ over NuAT. On healthy runs the guard statistic never came near its threshold, so the guarded number is the base method's own strength with the coin flip removed. The complete AutoAttack suite (the targeted FAB and square attacks in addition to the two APGD variants) finds nothing beyond APGD on our models, consistent with the absence of gradient masking. The CR attack, the benchmark's second strong-attack suite, preserves the ordering: ours $50.44{\pm}0.16$, ahead of NuAT ($50.11{\pm}0.19$) and LIET ($49.93{\pm}0.15$) and within $0.1$ of the multi-step PGD-AT-WA ($50.51{\pm}0.16$). The only better number in Table~\ref{tab:main} is PGD-AT-WA, a multi-step method at several times the cost. Table~\ref{tab:c10probe} (Appendix~\ref{app:seeds}) collects the threshold, control, and failure-mode probes.

\paragraph{CIFAR-100 and Tiny-ImageNet: the coupling weight flips.} \label{sec:nuc} The nuclear-norm term that helps on CIFAR-10 is toxic at the same strength on the harder datasets: at weight $4.0$ the base method collapses to $11.2$ on CIFAR-100 and $7.8$ on Tiny-ImageNet, the same failure as the published nuclear-norm method NuAT. Reducing the weight to $1.0$ repairs the setup, and the response on CIFAR-100 is an inverted U over the weight: $1.0$ gives $26.6$, the peak at $1.25$ with MSE weight $4.5$ gives $\mathbf{27.15{\pm}0.04}$ over four seeds, $1.5$ gives $27.1$, and $2.0$ falls back to $25.5$. At the peak we clear LIET ($26.64$) by $+0.51$ and the multi-step PGD-AT-WA ($26.90$, roughly twice our cost) by $+0.25$; the CR column shows the same margins ($27.20{\pm}0.07$, $+0.57$ over LIET and $+0.33$ over the multi-step reference). On Tiny-ImageNet the same re-balancing brings no further gain ($m_{\max}{=}5.0$ scores $20.69$ on its single seed, level with the $m_{\max}{=}2.5$ mean), so the setup stays at $\gamma_{\max}{=}1.0$, $m_{\max}{=}2.5$ and reaches $20.47{\pm}0.11$, $+0.59$ over LIET, the best fast-method number we are aware of on this dataset. The response is monotone on both sides of the flip, which identifies the nuclear weight, not the guard, as the load-bearing choice on these datasets. The guard never fires on either one, so here the contribution is the repair of the nuclear-norm family's known failure on harder datasets, with the guard carried as insurance. Per-seed values for every number in the table are listed in Appendix~\ref{app:seeds}.

\paragraph{Is the gain attack-specific?} No. Which samples a model solves under AutoAttack is strongly conserved across healthy methods (mean pairwise Kendall $\tau=0.82$), and our anchored models follow the same conserved pattern ($\tau=0.847$); the only models that rewrite the sample-difficulty spectrum are CO ones ($\tau=0.386$). The gain is therefore a genuine robustness gain, not an artifact of the evaluation attack. The check is worth reporting in its own right, since evaluation pipelines can silently favor some methods when test-side information leaks into training choices \cite{pan2025robustbench}; solving the same conserved set of hard samples is the opposite signature. Details in Appendix~\ref{app:spectrum}.

\section{Limitations} \label{sec:limits} (1)~Seed counts are small (4/4/3 per dataset). Zero CO in four seeds is consistent with a CO rate as high as $53\%$ (one-sided 95\% binomial bound); what we actually observe on healthy runs is that the guard statistic never approached its threshold, which is the operative claim. (2)~The per-dataset coupling weights ($\gamma_{\max}{=}4.0/1.25/1.0$) were found by small sweeps, not derived from the mechanism; the dose response is broad near the peak, but the weight is not universal across datasets. (3)~Not every failure of the base method is a validation-visible CO: one collapsed seed kept high validation PGD-10 while test robustness died (Appendix~\ref{app:seeds}), and no validation-side statistic we know sees that flavor. (4)~The propositions are informal, and the logit-margin construction is a sketch, not a proof of existence at network scale. (5)~Instrumentation perturbs the random streams: adding batch-level probes changes which seeds collapse, even though the dynamics for a given draw are deterministic.

\section{Conclusion} \label{sec:conclusion} Catastrophic overfitting is not an optimization failure in the usual sense: nothing about the training objective got worse. It is an escape along directions the training loss cannot see, by a model whose clean behavior quietly improves while its robustness disappears. Once measured, the geometry dictates the response: anchor the drifting weights to their own time-average, with a trigger that is itself sensitive to the drift. The anchor is attack-agnostic (no evaluation signal enters training), costs nothing beyond what the protocol already computes, and fails informatively wherever there is no robust region to anchor to.


{\small \bibliographystyle{plainnat} \bibliography{references} }

\appendix

\section{Protocol and measurement details} \label{app:proto} Training follows the benchmark \cite{pan2026benchmark} exactly: ResNet18 on CIFAR-10/100 and PreActResNet18 on Tiny-ImageNet; $\eps{=}8/255$ for training and evaluation; 100 epochs, OneCycle SGD, label smoothing $0.4$ \cite{szegedy2016ls}; a 49{,}000/1{,}000 train/validation split; per-epoch validation PGD-10 for checkpoint selection; the complete standard AutoAttack suite (apgd-ce, apgd-t, FAB-T, square) and the constraint-relaxation (CR) attack \cite{pan2025crattack}, both in the benchmark's released configuration on the full test set. The main-table and per-seed rows use the complete suite; the dose-probe rows keep the two-attack development variant, which on our models differs by at most $0.02$ (the targeted FAB and square attacks find no additional points on any of our checkpoints, consistent with the absence of gradient masking). A run of our method takes $1.05$ hours against $0.7$--$0.8$ for the strongest two-pass fast methods and about $2$h for PGD-AT. \emph{Measurement detail.} An EMA checkpoint evaluated with its own stale BatchNorm statistics scores at chance level, so every EMA evaluation in Sec.~\ref{sec:anatomy} recalibrates BatchNorm statistics on 4{,}096 training samples first; without this step the entire effect disappears. The recalibration is an analysis-time tool only and is not part of the deployed method. A \emph{CO event} is defined on the per-epoch validation curves so that it can be re-derived from any log: validation PGD-10 drops more than 30 points in one epoch, stays below $10\%$ for the following three epochs, while validation FGSM accuracy remains above $30\%$. The CIFAR-10 runs of Eq.~\eqref{eq:loss} additionally carry a small wrong-logit dispersion term (weight $0.3$) inherited from the method's development lineage; it is disabled on CIFAR-100/Tiny-ImageNet and plays no role in the anchoring mechanism. Odd steps add the same nuclear term inside the attack objective, following NuAT.

\section{Per-seed values} \label{app:seeds}

Tables~\ref{tab:c10probe}--\ref{tab:tiny} list the per-seed values behind every mean in the paper. Complete training logs, checkpoints, and the analysis outputs behind Sec.~\ref{sec:anatomy} ship with the supplementary material.

\begin{table}[t]
\centering
\caption{CIFAR-10: per-seed AutoAttack (\%) for our configurations and controls. Guard fires per seed in parentheses where nonzero.}
\label{tab:c10probe} \scriptsize \setlength{\tabcolsep}{2pt}
\resizebox{\linewidth}{!}{%
\begin{tabular}{lcc}
\toprule
Configuration & Per-seed AA (\%) & Note \\
\midrule
\multicolumn{3}{l}{\emph{ours}}\\
Base + guard, $\theta{=}8$ & 50.19/50.40/50.61/50.60 & fires 0/0/4/0 \\
guard $\theta{=}5$ & 50.72/50.43/50.57/50.63 & no change \\
guard $\theta{=}3$ & 50.42/49.90/50.54 & 2/5/3 spurious \\
\midrule
\multicolumn{3}{l}{\emph{threshold and trigger controls}}\\
guard $\theta{=}10$, same traj. & 7.94 & misses epoch 19 \\
guard $\theta{=}12$ & 50.40 / 7.84 & saves 1 of 2 seeds \\
guard, no cooldown (s2) & 50.18 & 15 consecutive fires \\
trigger on raw reading only & 7.54 & same draw, with budget \\
\phantom{trigger} without budget & 15.14 & selection artifact \\
\phantom{trigger} two readings & 7.53 & decision-identical \\
\midrule
\multicolumn{3}{l}{\emph{anchor controls}}\\
fixed-epoch reset & 49.99 & no detector \\
cont.\ pull $\eta_a{=}0.02$ & 50.45/50.29/50.39/7.14/50.11 & 1 of 5 fails \\
cont.\ pull $\eta_a{=}0.05$ & 8.15 & too strong \\
coupled-loss fallback & 7.05 & on fire \\
\midrule
\multicolumn{3}{l}{\emph{two-pass base (no clean branch), seed 0}}\\
per-example placement, $\pm$guard & 49.92 / 49.93 & \\
anchored at the init point & 49.85 & \\
straddle init $\alpha{=}0.75$ / $0.5$ & 49.84 / 40.23 & one CO, rescued \\
EMA-weights anchor & collapses & adv.\ acc.\ 0.94 by ep.\ 8 \\
high-weight base + guard & 7.13 / 7.80 / 5.07 & 3 seeds, no region \\
earlier two-pass ports & 48.57--49.84 & \\
\midrule
\multicolumn{3}{l}{\emph{decoupled victim (vanilla FGSM-AT)}}\\
continuous pull / guard & 0.00 / 38.34 & matches the own-EMA \\
guard, no cooldown/budget & 37.99 & benchmark rescue level \\
\bottomrule
\end{tabular}}
\end{table}

\begin{table}[t]
\centering
\caption{CIFAR-10: full test-set columns per model. The unguarded base either matches the guarded row (survivors) or returns a gradient-masked artifact (CO seeds: inflated clean, collapsed PGD-50 and AutoAttack).}
\label{tab:c10attack} \footnotesize \setlength{\tabcolsep}{4.5pt}
\begin{tabular}{llcccc}
\toprule
Model & Seed & Clean & PGD-10 & PGD-50 & AA \\
\midrule
Base + guard & 0 & 79.46 & 55.43 & 54.60 & 50.19 \\
 & 1 & 80.62 & 55.18 & 54.43 & 50.40 \\
 & 2 & 79.35 & 55.20 & 54.52 & 50.61 \\
 & 3 & 79.28 & 55.67 & 54.98 & 50.60 \\
\midrule
Base, no guard & 0 & 79.84 & --- & 54.94 & survived \\
 (seed status) & 3 & 79.28 & --- & 54.98 & survived \\
 & 1 & 92.58 & --- & 25.62 & 7.57 \\
 & 2 & 92.37 & --- & 24.61 & 7.48 \\
\bottomrule
\end{tabular}

\vspace{2pt} \raggedright\scriptsize PGD-20 for the guarded seeds: 54.74/54.58/54.65/55.05. Unguarded survivors score the same level as the guarded row under checkpoint selection restricted to healthy epochs; the two CO seeds select masked artifacts.
\end{table}

\begin{table}[t]
\centering
\caption{CIFAR-100: per-seed AutoAttack (\%) along the weight response. $\gamma_{\max}$ is the nuclear weight, $m_{\max}$ the MSE weight.}
\label{tab:c100} \scriptsize \setlength{\tabcolsep}{3pt}
\begin{tabular}{lcc}
\toprule
Setting & Per-seed AA (\%) & Mean \\
\midrule
$\gamma{=}4.0$ (one seed) & 11.20 & \\
$\gamma{=}0.5$ (one seed) & 22.44 & \\
$\gamma{=}1.0$, $m{=}2.5$ & 25.53/26.41/26.23/26.24 & 26.10$\pm$0.39 \\
no nuclear term (3 seeds) & 26.36/25.82/24.93 & \\
$m{=}3.5$ & 26.35/26.74/26.27/26.42 & 26.45 \\
$m{=}4.5$ & 26.61/26.35/26.66/26.64 & 26.57$\pm$0.14 \\
$m{=}5.0$ / $5.5$ (seed 0) & 26.44 / 25.98 & turnover above 4.5 \\
$\gamma{=}1.5$ (seed 0) & 27.12 & \\
$\gamma{=}2.0$ (seed 0) & 25.45 & past the peak \\
$\gamma{=}1.25$, $m{=}4.5$ (ours) & 27.20/27.13/27.13/27.13 & \textbf{27.15$\pm$0.04} \\
\midrule
\multicolumn{3}{l}{\emph{single-seed probes that did not move the result}}\\
Bernoulli-uniform init & 25.47 & \\
CE at the adversarial point, mix & 26.06 & \\
tail-only nuclear & 26.05 & \\
label smoothing $0.25$ & 25.35 & \\
CE mix + tail-only nuclear & 25.53 & \\
\midrule
$\theta{=}5$ / $\theta{=}3$ (seed 0) & 26.63 / 26.75 & 0 / 2 fires \\
\bottomrule
\end{tabular}
\end{table}

\begin{table}[t]
\centering
\caption{Tiny-ImageNet: per-seed AutoAttack (\%).}
\label{tab:tiny} \footnotesize
\begin{tabular}{lcc}
\toprule
Setting & Per-seed AA (\%) & Note \\
\midrule
$\gamma{=}1.0$, $m{=}2.5$ (ours) & 20.49/20.57/20.36 & 20.47$\pm$0.11 \\
$\gamma{=}4.0$ (one seed) & 7.80 & toxic \\
$m{=}5.0$ (one seed) & 20.69 & one guard fire \\
$\theta{=}5$ / $\theta{=}3$ (seed 0) & 20.42 / 20.40 & fire-free, level \\
\bottomrule
\end{tabular}
\end{table}

\section{Controls and failure modes} \label{sec:dose}

\paragraph{The rescue comes from the anchor, not the detector.} Three controls separate the mechanism from the bookkeeping. First, a detector-free control: reset at a fixed epoch (18), ignoring all signals. It rescues the same seed to $49.99$, essentially the guarded result. The detection signal contributes timing, not magic. (The fixed epoch presupposes knowledge of when that seed collapses, so this is a control, not a method.) Second, a strength control: an always-on anchor at $2.5\times$ the pull strength ($\eta_a{=}0.05$) destroys training ($8.15$). Only the blind component should ever be pulled; pulling always fights the optimizer itself. Third, falling back to the coupled loss when the guard fires instead of resetting also fails ($7.05$), confirming that it is the reset that helps, not any extra regularization. A re-nucleating draw (five CO events in 31 epochs) isolates the bookkeeping choices on a failure case: the raw-reading trigger is decision-identical to the two-reading version ($7.54$ versus $7.53$ on the same draw, the trajectories bit-identical until the budget binds), and without the fire budget checkpoint selection alone lifts the masked endpoint to $15.14$.

\paragraph{A continuous variant, and why we did not choose it.} \label{sec:trigger} The same idea can be made continuous: every $k{=}8$ steps, measure on the current batch the KL divergence $\Delta_t$ between the EMA's predictions and the adversarial predictions, and update
\begin{equation}
w_{t+1}=w_t-\eta_a\,\max(0,\Delta_t-\delta)\,(w_t-\wema_t).
\end{equation}
The pull is exactly zero while the two checkpoints agree and grows with the disagreement; it costs one extra forward pass per 8 steps. It failed on one of its five seeds ($7.14$): that seed's degradation was slow, and its trigger, the KL between clean and adversarial output distributions, stayed inside its dead zone the whole time. The reason is geometric and worth stating as a design principle: a trigger measured on the clean function inherits the clean function's blindness (Sec.~\ref{sec:def}) and can sleep through the very drift it is meant to catch. The guard's trigger is validation PGD, a robustness measurement rather than a clean one, and it fired and recovered the same seed. This is why the guard, not the continuous pull, is our method.

\paragraph{Where anchoring cannot help.} \label{sec:scope} An anchor needs a robust region to anchor to; the failures are as instructive as the successes. (i)~Vanilla FGSM training cannot be saved: its drift carries a persistent bias that drags the EMA along; anchoring yields $0.00$ (continuous pull) or $38.34$ (guard, merely matching the benchmark's own EMA-rescued FGSM-AT), and removing the guard's two constants does worse, not better, because it resets every epoch to an anchor that is itself sinking. The guard is not a plug-in cure for arbitrary fast methods; it protects coupled setups whose average stays put. (ii)~A high-weight coupled configuration at two-pass cost never establishes a stationary robust region in the first place (best validation PGD $36$--$43\%$); the guard fires, but the anchor it resets to is itself adrift, and all three seeds die. (iii)~Our base, with the clean branch in the graph, holds both properties at once: strong coupling and a stationary robust region. The third forward pass is not just cost; it is what makes the method anchorable.

\section{Additional analyses} \label{app:analyses}

\begin{figure}[t]
\centering \includegraphics[width=\linewidth]{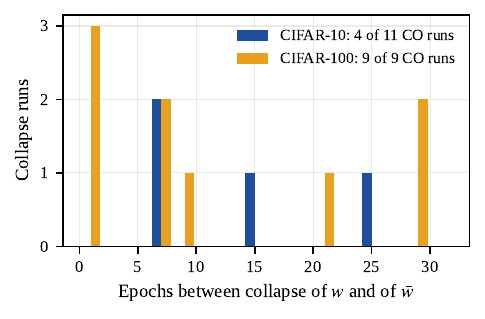}
\caption{The average collapses late, or never. Epochs between the raw CO and the EMA's own CO, over the CO runs recoverable from the released benchmark curves. Bars count only runs in which the EMA also collapses within 100 epochs: 4 of the 11 CIFAR-10 CO runs (lags $6$--$24$) and all 9 recoverable CIFAR-100 CO runs (lags $1$--$29$). In the remaining 7 CIFAR-10 CO runs the EMA never collapses within 100 epochs, so no finite lag exists.}
\label{fig:lag}
\end{figure}

\begin{figure}[t]
\centering \includegraphics[width=0.92\linewidth]{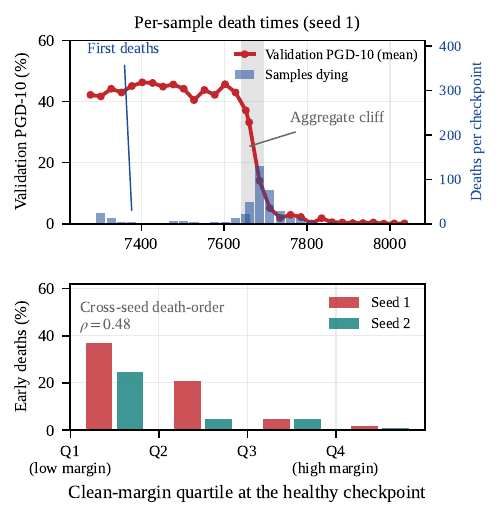}
\caption{CO in sample space. Top: validation PGD-10 (mean, left axis) and the number of samples dying per checkpoint (right axis) across the seed-1 window of Fig.~\ref{fig:micro}; a cohort of samples dies $100$ or more batches before the aggregate cliff. Bottom: fraction of each clean-margin quartile dying at least $100$ batches before the median, for two independently trained seeds; low-margin samples die first in both.}
\label{fig:death}
\end{figure}

\paragraph{Death-order statistics.} Early deaths have a smaller clean logit margin at the window's first checkpoint (median $1.02$ versus $1.62$; two-sided Mann--Whitney $p{=}9\times10^{-16}$; Spearman $-0.54$). The order is a property of the samples, not of one training path: an independently trained replica, instrumented on a fourfold coarser checkpoint grid, reproduces the same statistics ($p{=}2\times10^{-24}$, Spearman $-0.58$), its death order correlates with the first replica's at Spearman $\rho{=}0.48$ ($p{=}2\times10^{-21}$), and one replica's margins predict the other's death order ($-0.45$). Stratifying the trigger by margin does not fire earlier than the aggregate mean.

\paragraph{Precursor statistics.} At epoch resolution, variance and lag-1 autocorrelation statistics of the released curves fire in only 3 of 11 benchmark CO runs, and the FGSM$-$PGD gap is not a warning (15-point gaps occur in healthy runs). At batch resolution the ramp of Fig.~\ref{fig:micro} also carries the statistical signature of a system approaching a transition point: in the final 3{,}000 batches before the cliff the variance of the cosine rises by $150$--$280\times$ and its lag-1 autocorrelation climbs from $0.69$--$0.73$ to $0.88$ in both instrumented seeds, the pattern known as critical slowing down.

\paragraph{Drift anatomy.} The weight difference across the seed-1 CO pair concentrates moderately in late blocks (energy Gini $0.79$; the eight largest tensors, all convolutions in stages 3--4, hold $58\%$ of the energy) but is not surgically localizable: rolling back the top-$20\%$-energy tensors to the healthy checkpoint recovers only $0.043\to0.133$ of the $0.409$ validation PGD-10, and a random equal-size rollback is harmful ($0.010$). The CO load is diffuse across shared features, which is why the guard rolls back the whole weight vector rather than a submodule.

\paragraph{Two flavors of validation blindness.} On saved checkpoints we compared the protocol's selection statistic (validation PGD-10 on 1{,}000 held-out images) with the attack-direction cosine used in Fig.~\ref{fig:micro}. A collapsed checkpoint selected by validation PGD (the hard draw of the no-budget ablation: validation PGD $0.432$, AutoAttack $7.5$) is caught by geometry: its FGSM accuracy stays high ($0.740$) while the cosine collapses to $0.055$, so multiplying the selection score by the cosine costs nothing and rejects it. A second collapsed seed shows the opposite flavor: validation PGD-10 stays high ($0.592$) and the cosine healthy ($0.854$) to the end of training while test PGD-50 falls to $25.6$ and clean accuracy inflates to $92.6\%$; the validation distribution itself has been memorized. No validation-side statistic can see this flavor, which is why the guard's rescue is limited to CO events that are visible in validation robustness.

\section{Which samples do the methods solve?} \label{app:spectrum}

We score per-sample AutoAttack survival (500-sample common validation subset) and compare orderings across models by Kendall $\tau$ (Table~\ref{tab:tau}). Healthy methods, benchmark and ours alike, agree strongly on which samples are hard (pairwise $\tau$ $0.79$--$0.91$, mean $0.82$); a clean-difficulty baseline reaches $0.79$, so much of the agreement is generic difficulty. PGD-20 with 5 restarts reproduces the AutoAttack ordering almost perfectly (per-model agreement $0.984$), so the conservation is not an AutoAttack artifact. The only spectrum rewriters are failure modes: a collapsed seed ($\tau=0.386$) and a logit-surgery variant ($0.624$). The strongest surviving base variant (AA $50.67$) does not move the spectrum ($0.847$): high mean accuracy here is earned on the same difficulty structure as every other healthy method, not by gaming a particular attack.

\begin{table}[h]
\centering
\caption{Kendall $\tau$ of per-sample AutoAttack survival. All seven measured values are listed.}
\label{tab:tau} \small
\begin{tabular}{lc}
\toprule
Quantity & $\tau$ \\
\midrule
Pairwise mean over 16 healthy models & 0.82 \\
PGD-20 (5 restarts), pairwise & 0.870 \\
PGD-20 vs.\ AutoAttack ordering, per model & 0.984 \\
Clean-difficulty baseline & 0.79 \\
Collapsed seed & 0.386 \\
Logit-surgery variant & 0.624 \\
Strongest surviving base variant (AA $50.67$) & 0.847 \\
\bottomrule
\end{tabular}
\end{table}

\end{document}